\documentclass{article}

\usepackage{latexsym}
\usepackage{enumitem}
\usepackage{float}
\usepackage{booktabs}
\usepackage{multirow, makecell}
\usepackage{enumitem}
\usepackage{subfigure}
\usepackage{graphicx}

\usepackage{microtype}
\usepackage{caption}
\usepackage{makecell}
\usepackage{amsmath}

\newlength{\mysize}

\usepackage{hyperref}

\usepackage[accepted]{icml2023}

\usepackage{amsmath}
\usepackage{amssymb}
\usepackage{mathtools}
\usepackage{amsthm}
\usepackage{multirow}
\usepackage{enumitem}

\usepackage[capitalize,noabbrev]{cleveref}

\begin{document}

\icmltitlerunning{Large Language Models Struggle to Learn Long-Tail Knowledge}
\twocolumn[
\icmltitle{Large Language Models Struggle to Learn Long-Tail Knowledge}

\begin{icmlauthorlist}
\icmlauthor{Nikhil Kandpal}{unc}
\icmlauthor{Haikang Deng}{unc}
\icmlauthor{Adam Roberts}{google}
\icmlauthor{Eric Wallace}{berkeley}
\icmlauthor{Colin Raffel}{unc}
\end{icmlauthorlist}

\icmlaffiliation{unc}{UNC Chapel Hill}
\icmlaffiliation{google}{Google Research}
\icmlaffiliation{berkeley}{UC Berkeley}

\icmlcorrespondingauthor{Nikhil Kandpal}{nkandpa2@cs.unc.edu}

\icmlkeywords{Language Models, NLP, Stealing, Security}

\vskip 0.3in
]

\printAffiliationsAndNotice{}

\begin{abstract}
The Internet contains a wealth of knowledge---from the birthdays of historical figures to tutorials on how to code---all of which may be learned by language models.
However, while certain pieces of information are ubiquitous on the web, others appear extremely rarely.
In this paper, we study the relationship between the knowledge memorized by large language models and the information in pre-training datasets scraped from the web.
In particular, we show that a language model's ability to answer a fact-based question relates to how many documents associated with that question were seen during pre-training.
We identify these relevant documents by entity linking pre-training datasets and counting documents that contain the same entities as a given question-answer pair. 
Our results demonstrate strong correlational and causal relationships between accuracy and relevant document count for numerous question answering datasets (e.g., TriviaQA), pre-training corpora (e.g., ROOTS), and model sizes (e.g., 176B parameters).
Moreover, while larger models are better at learning long-tail knowledge, we estimate that today's models must be scaled by many orders of magnitude to reach competitive QA performance on questions with little support in the pre-training data.
Finally, we show that retrieval-augmentation can reduce the dependence on relevant pre-training information, presenting a promising approach for capturing the long-tail.
\end{abstract}
\section{Introduction}

\begin{figure}[t]
\centering
\includegraphics[trim={0.1cm, 0.1cm, 0.1cm, 0.1cm}, clip, width=\linewidth]{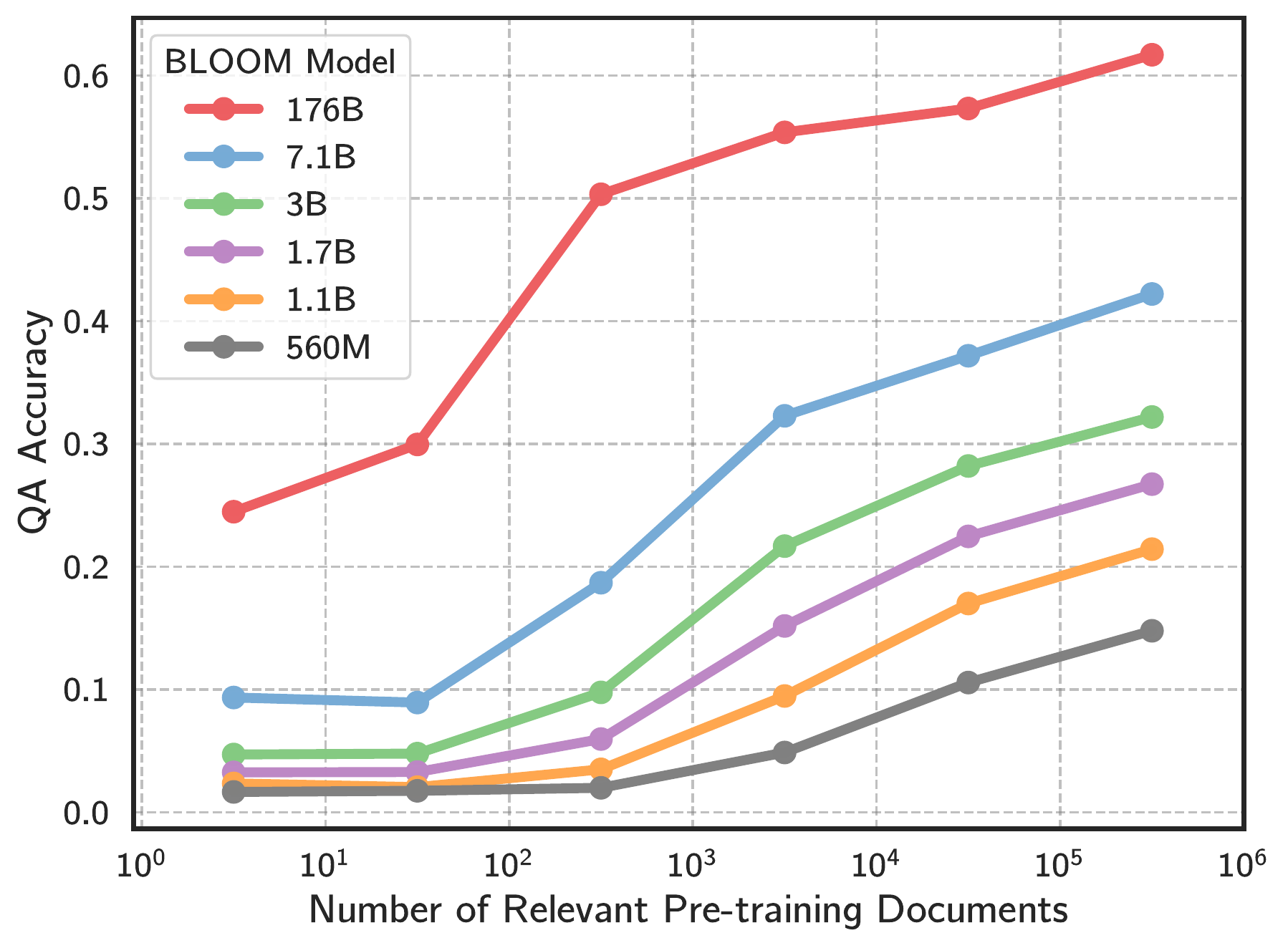}
\vspace{-0.6cm}
\caption{Language models struggle to capture the long-tail of information on the web. Above, we plot accuracy for the BLOOM model family on TriviaQA as a function of how many documents in the model's pre-training data are relevant to each question.}
\label{fig:bloom_tqa}
\end{figure}

Large language models (LLMs) trained on text from the Internet capture many facts about the world, ranging from well-known factoids to esoteric domain-specific information.
These models implicitly store knowledge in their parameters \cite{petroni2019language,roberts2020much}, and given the scale of today's pre-training datasets and LLMs, one would hope that they can learn a huge amount of information from web-sourced text.
However, not all of the knowledge on the Internet appears equally often---there is a long-tail of information that appears rarely or only once.

\begin{figure*}[t]
\centering
\includegraphics[trim={0cm, 4cm, 3.85cm, 0.1cm}, clip, width=0.98\linewidth, page=1]{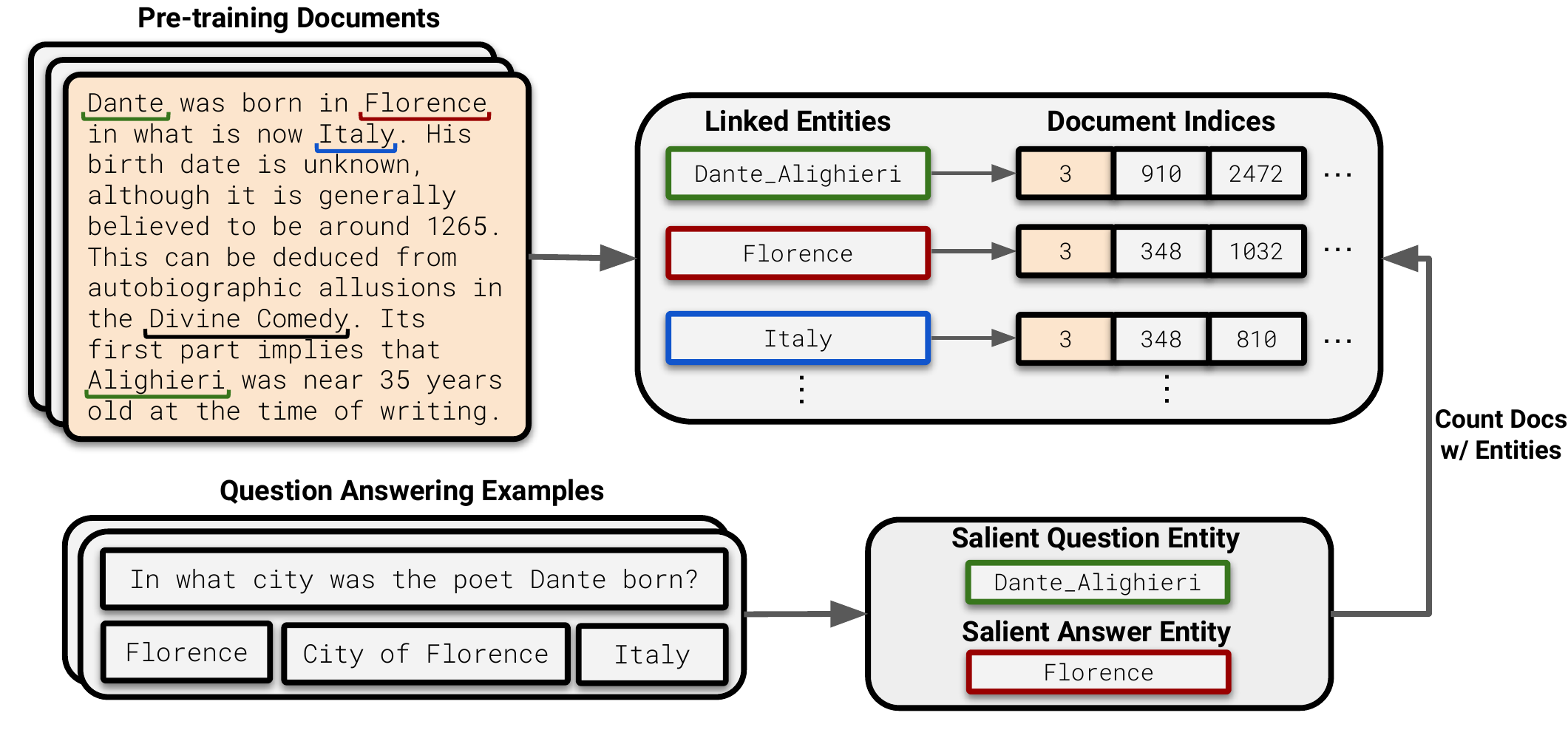}
\vspace{-0.45cm}
\caption{In our document counting pipeline, we first run entity linking on large pre-training datasets (\textit{top left}) and store the set of the document indices in which each entity appears (\textit{top right}).
We then entity link downstream QA pairs and extract the salient question and answer entities (\textit{bottom}). Finally, for each question we count the number of documents in which the question and answer entities co-occur.}
\label{fig:pipeline_diagram}
\end{figure*}

In this work, we explore the relationship between the knowledge learned by an LLM and the information in its pre-training dataset.
Specifically, we study how an LLM's ability to answer a question relates to how many documents associated with that question were seen during pre-training. 
We focus on factoid QA datasets~\cite{joshi2017triviaqa,kwiatkowski2019natural}, which lets us ground question-answer pairs into concrete subject-object co-occurrences. As an example, for the QA pair (\textit{In what city was the poet Dante born?}, \textit{Florence}), we consider documents where the entities \texttt{Dante} and \texttt{Florence} co-occur as highly relevant. To identify these entity co-occurrences we apply a highly-parallelized entity linking pipeline to trillions of tokens from datasets such as C4~\cite{raffel2020exploring}, The Pile~\cite{gao2020pile}, ROOTS~\cite{roots}, OpenWebText~\cite{Gokaslan2019OpenWeb}, and Wikipedia.

We observe a strong correlation between an LM's ability to answer a question and the number of pre-training documents relevant to that question for numerous QA datasets, pre-training datasets, and model sizes (e.g., Figure~\ref{fig:bloom_tqa}).
For example, the accuracy of BLOOM-176B~\cite{bloom} jumps from 25\% to above 55\% when the number of relevant pre-training documents increases from $10^1$ to $10^4$. 

We also conduct a counterfactual re-training experiment, where we train a 4.8B-parameter LM with and without certain documents. Model accuracy drops significantly on questions whose relevant documents were removed, which validates our entity linking pipeline and shows that the observed correlational trends are likely causal in nature.

Finally, we analyze ways to better capture knowledge that rarely appears in the pre-training data: model scaling and retrieval-augmentation.
For model scaling, we find a strong log-linear relationship between parameter count and QA accuracy. These trends show that while scaling up LMs improves knowledge learning, models would need to be scaled dramatically (e.g., to one quadrillion parameters) to achieve competitive QA accuracy on long-tail questions. Retrieval-augmented systems are more promising---when a retriever succeeds in finding a relevant document, it reduces an LLM's need to have a large amount of relevant pre-training text. Nevertheless, retrieval systems themselves still exhibit a mild dependence on relevant document count. 

Overall, our work is one of the first to study how LLM knowledge is influenced by pre-training data. To enable future research, we release our code as well as the entity data for ROOTS, The Pile, C4, OpenWebText, and Wikipedia at \url{https://github.com/nkandpa2/long_tail_knowledge}.
\section{Identifying Relevant Pre-training Data}\label{sec:method}

\paragraph{Background and Research Question} Numerous NLP tasks are \textit{knowledge-intensive}: they require recalling and synthesizing facts from a knowledge source (e.g.\ Wikipedia or the web). Results on knowledge-intensive tasks have been dramatically improved using LLMs, as these models have been shown to leverage the vast amounts of knowledge they learn from their pre-training corpora~\cite{roberts2020much,petroni2019language,de2020autoregressive}. However, it remains unclear as to \textit{what kind} of knowledge LMs actually capture---for example, do they simply learn ``easy'' facts that frequently appear in their pre-training data?

We study this question using closed-book QA evaluations~\cite{roberts2020much} of LLMs in the few-shot setting~\cite{brown2020language}. Models are prompted with in-context training examples (QA pairs) and a test question without any relevant background text.
The goal of our work is to investigate the relationship between an LM's ability to answer a question and the number of times information relevant to that question appears in the pre-training data. 

\paragraph{Our Approach} The key challenge is to efficiently identify all of the documents that are relevant to a particular QA pair in pre-training datasets that are hundreds of gigabytes in size. 
To tackle this, we begin by identifying the salient entities that are contained in a question and its set of ground-truth answer aliases.
We then identify relevant pre-training documents by searching for instances where the salient question entity and the answer entity co-occur. 

For example, consider the question \textit{In what city was the poet Dante born?} with the valid answers \textit{Florence}, \textit{City of Florence}, and \textit{Italy} (e.g., Figure~\ref{fig:pipeline_diagram}). We extract the salient question and answer entities, \texttt{Dante\_Alighieri} and \texttt{Florence}, and count the documents that contain both entities. 

Our approach is motivated by \citet{trex}, who show that when only the subject and object of a subject-object-relation triple co-occur in text, the resulting triple is often also present. In addition, we conduct human studies that show our document counting pipeline selects relevant documents a majority of the time (Section~\ref{subsec:human_eval}). Moreover, we further validate our pipeline by training an LM without certain relevant documents and showing that this reduces accuracy on the associated questions (Section~\ref{subsec:counterfactual}). Based on these findings, we refer to documents that contain the salient question and answer entities as \textbf{relevant documents}. 

To apply the above method, we must entity link massive pre-training corpora, as well as downstream QA datasets. We accomplish this by building a parallelized pipeline for entity linking (Section~\ref{subsec:pretraining_analysis}), which we then customize for downstream QA datasets (Section~\ref{subsec:qa_analysis}).

\subsection{Entity Linking Pre-training Data}\label{subsec:pretraining_analysis}

We perform entity linking at scale using a massively distributed run of the DBpedia Spotlight Entity Linker~\cite{dbpedia}, which uses traditional entity linking methods to link entities to DBpedia or Wikidata IDs. 
We entity link the following pre-training datasets, which were chosen based on their use in the LLMs we consider:
\begin{itemize}[leftmargin=*,itemsep=1mm]
    \item \textbf{The Pile:} an 825GB dataset that contains a mix of 22 different primarily English-language sources~\cite{gao2020pile}.
    \item \textbf{ROOTS (En):} the 490GB English subset of the ROOTS corpus~\cite{roots}. Note that we do not study if models trained on the non-English subsets of ROOTS are able to leverage cross-lingual factual knowledge.
    \item \textbf{C4:} a 305GB English corpus that was collected by filtering CommonCrawl~\cite{raffel2020exploring}.
    \item \textbf{OpenWebText:} a 39GB English corpus that contains the text of web pages that were linked on the website Reddit~\cite{Gokaslan2019OpenWeb}.
    \item \textbf{Wikipedia:} a text dump of December 2018 Wikipedia articles from \citet{lee2019latent}, a standard corpus for evaluating open-domain QA systems (e.g.\ \citealt{karpukhin2020dense,lewis2020retrieval,guu2020retrieval}).
\end{itemize}

For each document in these pre-training datasets, we record the linked entities in a data structure that enables quickly counting individual entity occurrences and entity co-occurrences. This pipeline took approximately 3 weeks to entity link 2.1TB of data on a 128-CPU-core machine. 

\subsection{Finding Entity Pairs in QA Data}\label{subsec:qa_analysis}

We next entity link two standard open-domain QA datasets: Natural Questions~\cite{kwiatkowski2019natural} and TriviaQA~\cite{joshi2017triviaqa}.
To expand our sample sizes, we use both the training and validation data, except for a small set of examples used for few-shot learning prompts.

We first run the DBPedia entity linker on each example. Because there can be multiple annotated answers for a single example, we concatenate the question and all valid answers, as this enabled more accurate entity linking. We use the most common entity found in the set of ground truth answers as the salient answer entity. We then iterate over all entities found in the question and select the entity that co-occurs the most with the salient answer entity in the pre-training data. In cases where no entity is found in the question, answer, or both, we discard the example.
If the resulting number of relevant documents is zero, we discard the example, as this is likely due to an entity linking error.

\subsection{Human Evaluation of Document Counting Pipeline}\label{subsec:human_eval}

Here, we conduct a human evaluation of our document identification pipeline. Note that a document can vary in the extent to which it is ``relevant'' to a particular QA pair.
For instance, consider the QA pair (\textit{William Van Allan designed which New York building---the tallest brick building in the world in 1930?}, \textit{Chrysler Building}). The documents that we identify as relevant may (1) contain enough information to correctly answer the question, (2) contain information relevant to the question but \textit{not enough} to correctly answer it, or (3) contain no relevant information. 
For example, a document that mentions that the Chrysler building was designed by William Van Allan, but not that it was the tallest brick building in 1930, would fall into the second category. 

We randomly sample 300 QA pairs from TriviaQA and selected one of their relevant documents at random. We then manually labeled the documents into one of the three categories: 33\% of documents contained enough information to answer the question and an additional 27\% of contained some relevant information. Thus, our pipeline has $\sim$60\% precision at identifying relevant documents for TriviaQA.

Our pipeline is imperfect as (1) the entity linker sometimes mis-identifies entities and (2) not all documents containing the salient question and answer entity are relevant. However, when applied at the scale of large-scale pre-training datasets, this pipeline is efficient and achieves enough precision and recall to observe correlational (Section~\ref{subsec:correlation}) and causal (Section~\ref{subsec:counterfactual}) relationships to QA performance.
\section{LM Accuracy Depends on Relevant Document Count}

\begin{figure}
\centering
\includegraphics[width=\linewidth]{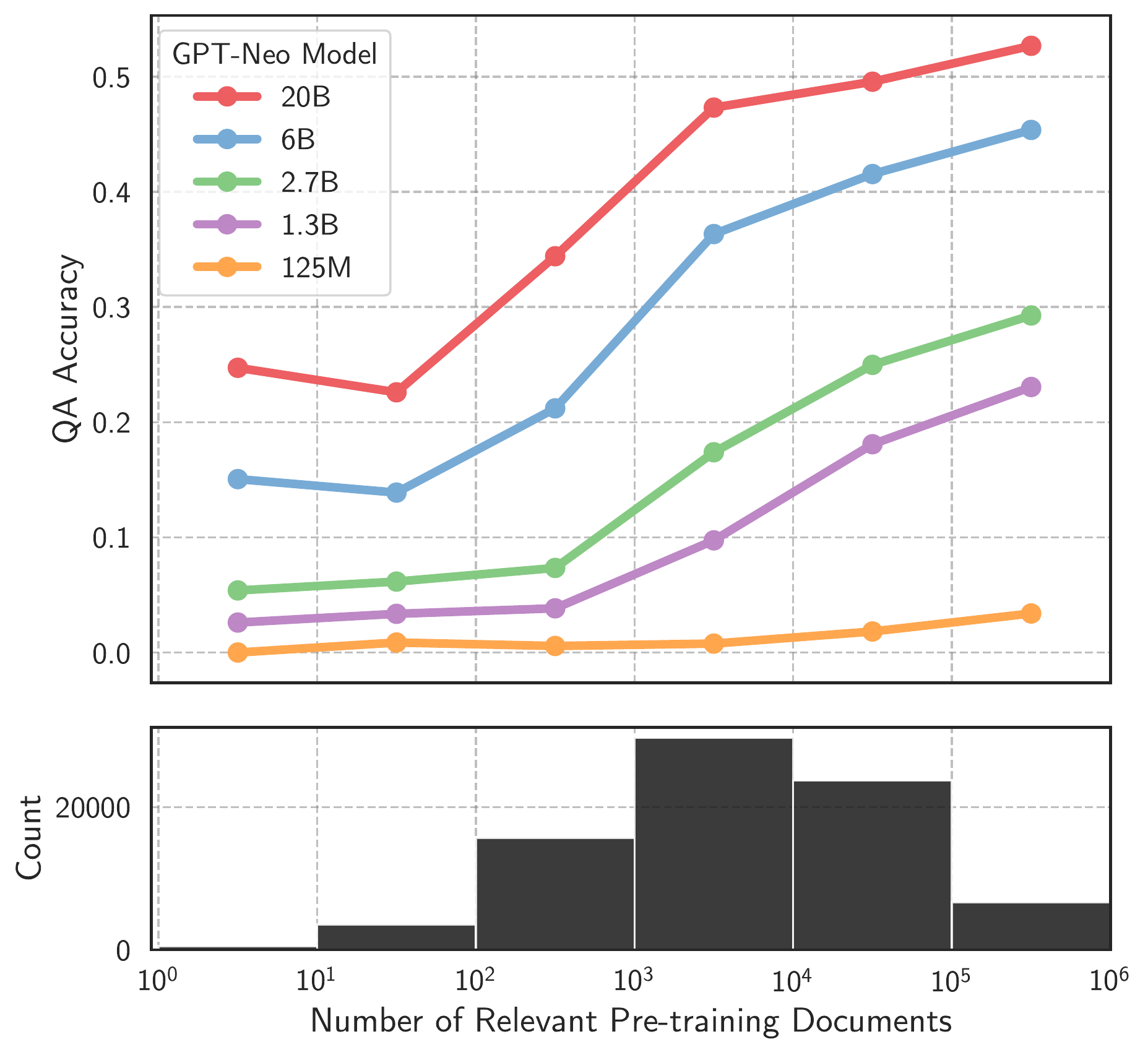}
\vspace{-0.7cm}
\caption{We plot accuracy on TriviaQA versus relevant document count for GPT-Neo. The trends match those seen for BLOOM (Figure~\ref{fig:bloom_tqa}). We also include a histogram that shows how many QA examples fall into each bucket; TriviaQA often asks about knowledge represented $10^2$ to $10^5$ times in the pre-training data.}
\label{fig:gpt_neo_tqa}
\end{figure}

\begin{figure}[htp]
\centering
\includegraphics[width=\linewidth]{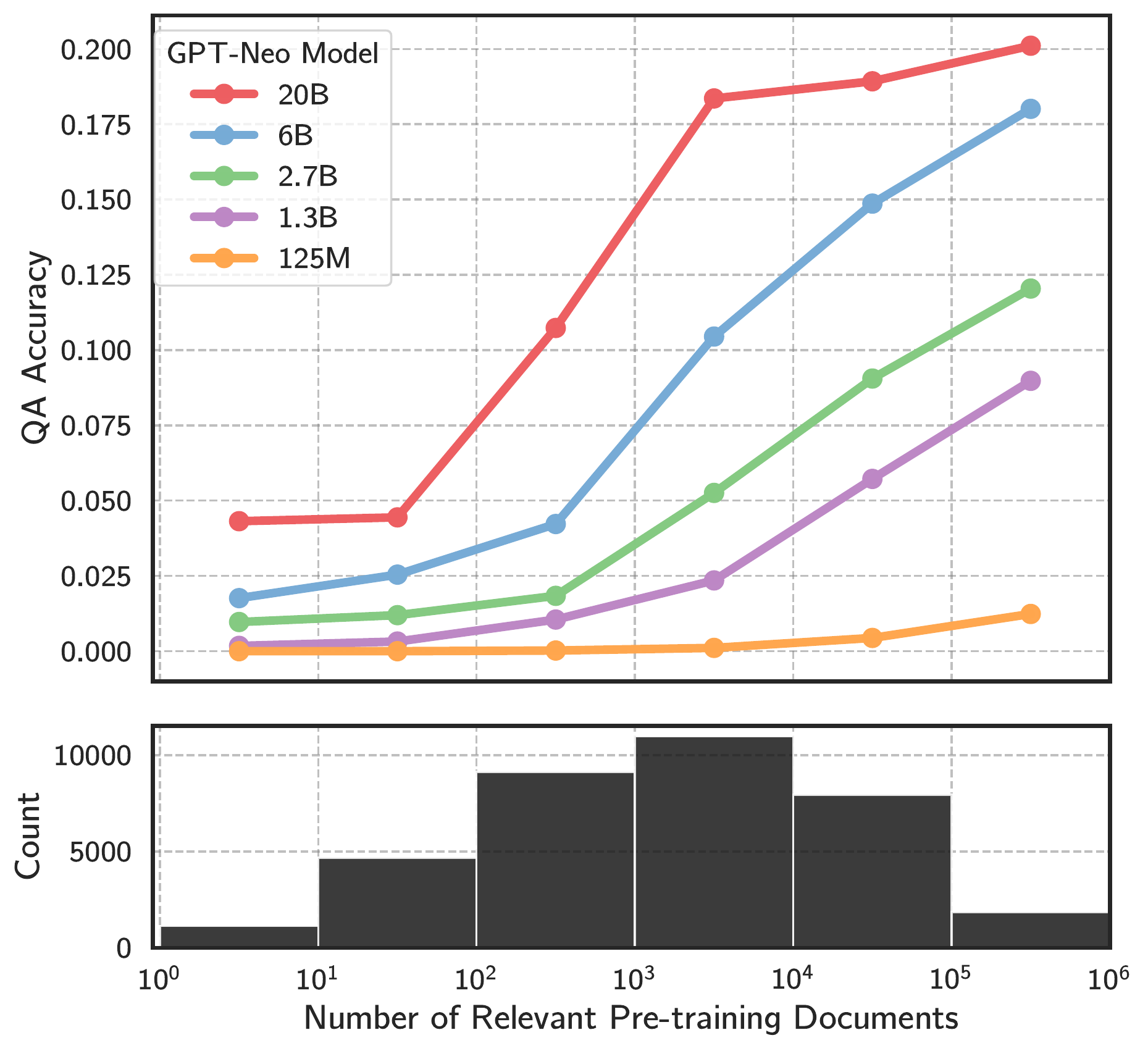}
  \label{fig:gpt_neo_nq} 
\vspace{-0.7cm}
\caption{We plot accuracy on Natural Questions versus relevant document count for GPT-Neo. The trends match those in TriviaQA---model accuracy is highly dependent on fact count.}
\label{fig:neo_nq}
\end{figure}

In this section, we measure the relationship between an LLM's ability to answer a question and the number of relevant documents in the pre-training corpus.
We use popular Transformer decoder-only LMs~\cite{vaswani2017attention} that span three orders of magnitude in size:
\begin{itemize}[leftmargin=*,itemsep=1mm]
    \item \textbf{GPT-Neo:} The GPT-Neo, GPT-NeoX, and GPT-J LMs trained by EleutherAI on the Pile~\cite{gao2020pile} that range in size from 125M to 20B parameters~\cite{gpt-neo,gpt-j,black2022gpt}. We refer to these models collectively as GPT-Neo models. 
    \item \textbf{BLOOM:} Models trained by the BigScience initiative on the ROOTS dataset~\cite{bloom}. The BLOOM models are multi-lingual; we analyze their English performance only. The models range in size from 560M to 176B parameters.
    \item \textbf{GPT-3:} Models trained by OpenAI that range in size from $\approx$350M (Ada) to $\approx$175B parameters (Davinci). Since the pre-training data for these models is not public, we estimate relevant document counts by scaling up the counts from OpenWebText to simulate if the dataset was the same size as GPT-3's pre-training data. 
    We recognize that there is uncertainty around these models' pre-training data, their exact sizes, and whether they have been fine-tuned. We therefore report these results in the Appendix for readers to interpret with these sources of error in mind. 
\end{itemize}
We use these LMs because (with the exception of GPT-3) they are the largest open-source models for which the pre-training data is publicly available.
We focus on 4-shot evaluation, although we found that other amounts of in-context training examples produced similar trends. We use simple prompts consisting of templates of
the form 
\begin{align*}
    &\texttt{Q: [In-Context Question 1]} \\ 
    &\texttt{A: [In-Context Answer 1]} \\
    &\hspace{5.5em} \vdots \\
    &\texttt{Q: [In-Context Question $n$]} \\ 
    &\texttt{A: [In-Context Answer $n$]} \\
    &\texttt{Q: [Test Question]}
\end{align*}
We generate answers by greedy decoding until the models generate a newline character, and we evaluate answers using the standard Exatch Match (EM) metric against the ground-truth answer set \cite{rajpurkar2016squad}.

\subsection{Correlational Analysis}\label{subsec:correlation}

We first evaluate the BLOOM and GPT-Neo model families on TriviaQA and plot their QA accuracies versus the number of relevant documents in Figures~\ref{fig:bloom_tqa} and \ref{fig:gpt_neo_tqa}.
For improved readability, we average the accuracy for QA pairs using log-spaced bins (e.g., the accuracy for all questions with 1 to 10 relevant documents, 10 to 100 relevant documents, etc.).
Below each plot, we also include a histogram that shows how many QA examples fall into each bin. We trim the plots when the bins contain fewer than 500 QA examples to avoid reporting accuracies for small sample sizes.

There is a strong correlation between question answering accuracy and relevant document count for all tested models.
Correspondingly, when the number of relevant documents is low, models are quite inaccurate, e.g., the accuracy of BLOOM-176B jumps from 25\% to above 55\% when the number relevant documents increases from $10^1$ to $10^4$.
Model size is also a major factor in knowledge learning: as the number of model parameters is increased, the QA performance substantially improves. For example, BLOOM-176B has over 4$\times$ higher accuracy than BLOOM-560M on TriviaQA questions with more than $10^5$ relevant documents.

\begin{figure}[t]
\centering
\includegraphics[trim={0.0cm 0.0cm 0.0cm 0.0cm},clip,width=\linewidth]{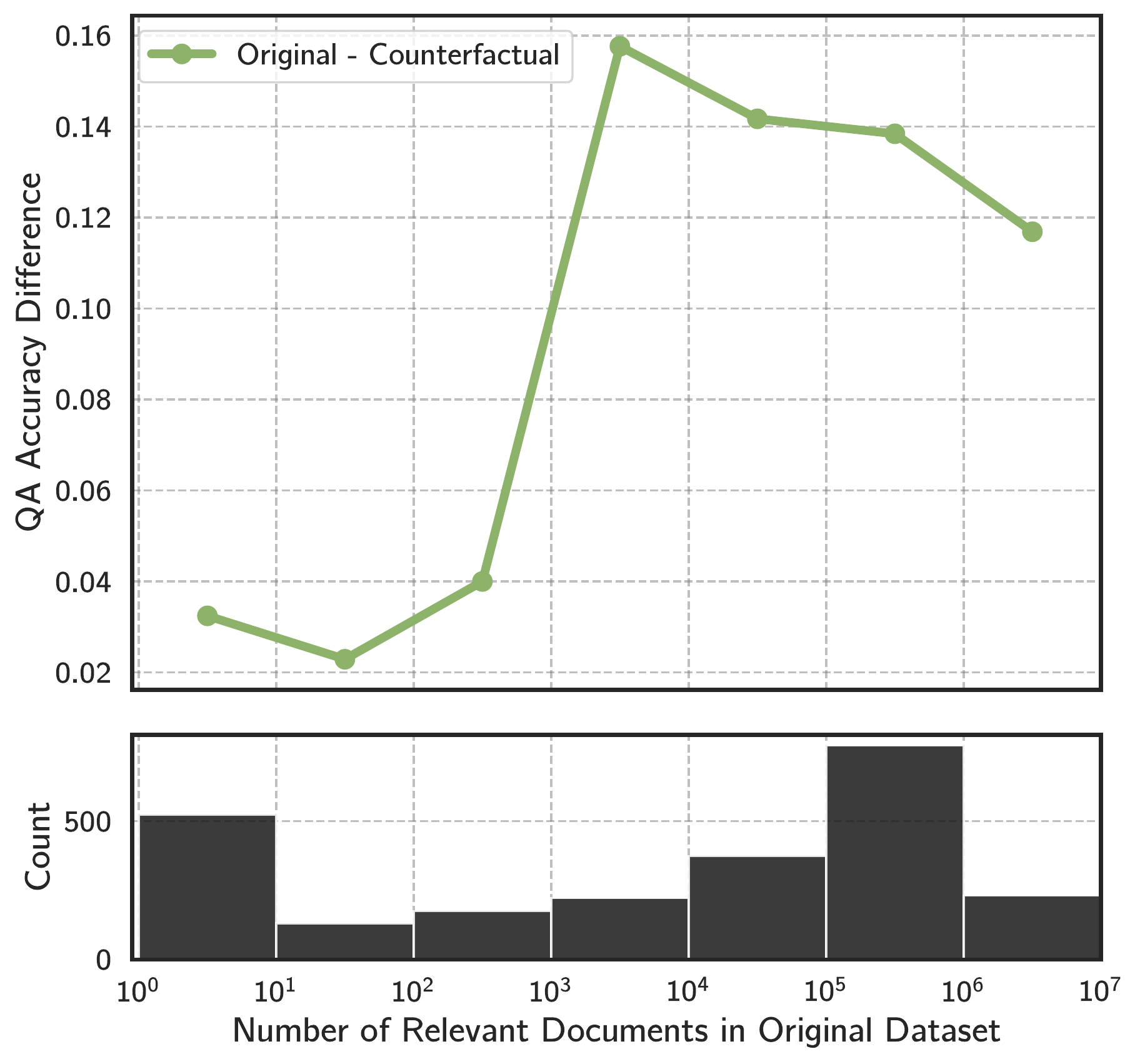}
\vspace{-0.7cm}
\caption{We run a counterfactual experiment, where we re-train an LM without certain documents. We take TriviaQA questions with different document counts and delete all of their relevant pre-training documents. The difference in accuracy between the original model and the re-trained LM (\textit{counterfactual}) is high when the original number of relevant documents is large. }
\label{fig:counterfactual}
\end{figure}

We repeat this experiment using the Natural Questions QA dataset and find similar trends for all model families (see Figure \ref{fig:neo_nq} for GPT-Neo, and Figures \ref{fig:nq} and \ref{fig:gpt3} in the Appendix for BLOOM and GPT-3 results).

\paragraph{Simpler Methods for Identifying Relevant Documents Are Less Effective} 
In the experiments above, we identify relevant documents by searching for co-occurrences of salient question and answer entities. To evaluate whether this process is necessary, we compare against two baseline document identification methods: counting documents that contain the salient question entity and counting documents that contain the salient answer entity (as done in \citealt{petroni2019language}). 

We show in Figure~\ref{fig:question_answer_only} that all three document identification methods are correlated with QA accuracy. However, when only considering QA examples where the question and answer entities co-occur few ($<5$) times, the two baseline methods no longer correlate with QA accuracy. This indicates that counting documents with just the answer entity or question entity alone is insufficient for explaining why LMs are able to answer certain questions.
This validates our definition of relevant documents as those that contain both the question entity and answer entity.

\paragraph{Humans Show Different Trends Than LMs} An alternate explanation for our results is that questions with lower document counts are simply ``harder'', which causes the drop in model performance. We show that this is not the case by measuring human accuracy on Natural Questions. We use a leave-one-annotator-out metric, where we take questions that are labeled by 5 different human raters (all of whom can see the necessary background text), hold out one of the raters, and use the other four as the ground-truth answer set.
We plot the human accuracy versus relevant document count in the top of Figure~\ref{fig:gold_augmented_fact_count}. 
Human accuracy is actually highest for the questions with \textit{few} relevant documents, the opposite trend of models.
We hypothesize that humans are better on questions with few relevant documents because (1) questions about rarer facts are more likely to be simple factoids compared to common entities, and (2) the Wikipedia documents are that are provided to the annotators are shorter for rarer entities, which makes reading comprehension easier and increases inner-annotator agreement.
\subsection{Causal Analysis via Re-training}\label{subsec:counterfactual}

Our results thus far are correlational in nature: there may be unknown confounds that explain them away, i.e., the rarer questions are more difficult for LMs for other reasons. Here we establish a causal relationship by removing certain documents in the training data and re-training the LM.

We first train a baseline 4.8 billion parameter LM on C4, following the setup from \citet{wang2022language}. We then measure the effect of deleting certain documents from the training set. For each log-scaled bin of relevant document count (e.g., $10^0$ to $10^1$ relevant documents, $10^1$ to $10^2$, ...) we sample 100 questions from Trivia QA and remove all relevant documents for those questions in C4. In total, this removes about 30\% of C4. Finally, we train a ``counterfactual'' LM on this modified pre-training dataset and compare its performance to the baseline model.
For both the baseline model and the counterfactual model, we train for a single epoch.
Note that the counterfactual model was trained for 30\% fewer steps, which makes it slightly worse in performance overall. To account for this, we only study the performance on questions whose relevant documents were removed.

We show the difference in performance between the two LMs on questions whose documents were removed in Figure~\ref{fig:counterfactual}.
For questions with few relevant documents in the original C4 dataset, performance is poor for both the baseline and the counterfactual LM, i.e., their performance difference is small.
However, for questions with many relevant documents, performance is significantly worse for the counterfactual LM.
This suggests a \textit{causal} link between the number of relevant documents and QA performance.

\begin{figure}[t]
\centering
\includegraphics[width=\linewidth]{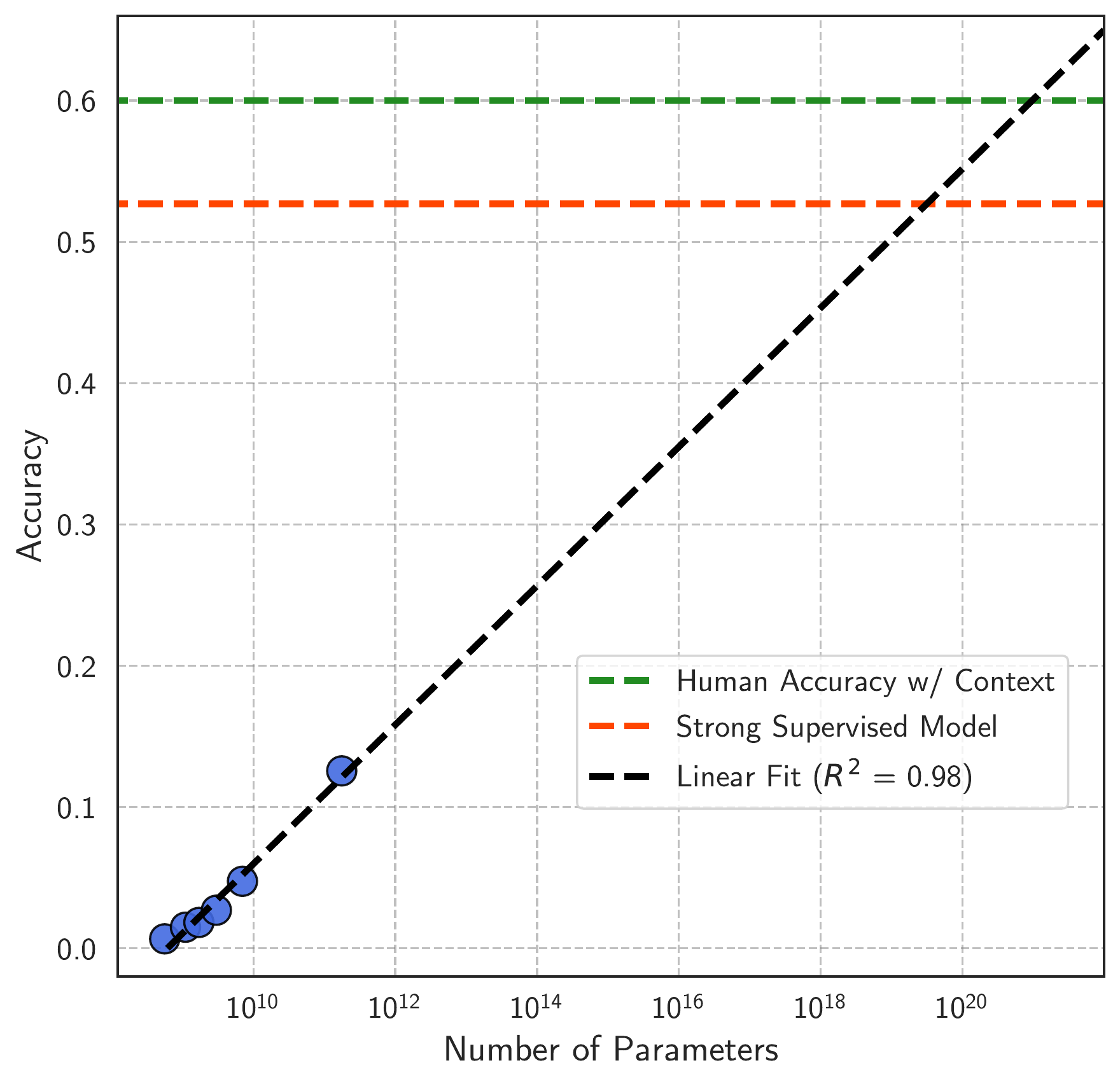}
\vspace{-0.8cm}
\caption{\emph{Scaling trends for fact learning}. We plot BLOOM accuracy on rare instances from Natural Questions ($<100$ relevant docs) as a function of the log of the model size. Extrapolating from the empirical line of best fit---which approximates the trend well at $R^2 = 0.98$---implies that immensely large models would be necessary to get high accuracy.}
\label{fig:bloom_nq_scale}
\end{figure}

\section{Methods to Improve Rare Fact Learning}

Thus far, we showed that LLMs have a strong dependence on relevant document count. Here, we investigate methods to mitigate this dependence: increasing data scale, increasing model scale, and adding an auxiliary retrieval module.

\subsection{Can We Scale Up Datasets?}\label{subsec:data_scaling}

Today's largest LLMs are pre-trained on hundreds of billions of tokens. One na\"ive approach for improving accuracy on questions about less-prevalent knowledge is to collect larger quantities of data. Our results suggest that this would not significantly improve accuracy as scaling datasets by moderate factors (e.g., 5x) usually results in small accuracy gains.
An alternative idea would be to increase the diversity of the pre-training data. However, we also believe this would provide minimal benefit because many data sources are surprisingly correlated . Although each of the pre-training datasets considered were collected independently, the amount of supporting information they provide for different TriviaQA examples is highly consistent as seen by the rank correlations between their relevant document counts in Table~\ref{tab:rank_correlation}.

\begin{table}[] 
\begin{tabular}{l|ccccc}
      & \multicolumn{1}{l}{ROOTS} & \multicolumn{1}{l}{Pile} & \multicolumn{1}{l}{C4} & \multicolumn{1}{l}{OWT} & \multicolumn{1}{l}{Wiki} \\ \hline
ROOTS & -                         & 0.97                     & 0.97                   & 0.94                    & 0.87                     \\
Pile  & -                         & -                        & 0.95                   & 0.96                    & 0.87                     \\
C4    & -                         & -                        & -                      & 0.96                    & 0.90                     \\
OWT   & -                         & -                        & -                      & -                       & 0.91                     \\
Wiki  & -                         & -                        & -                      & -                       & -      
\end{tabular}
\caption{Spearman rank correlations of the relevant document counts for TriviaQA examples in The Pile, ROOTS, C4, OpenWebText, and Wikipedia. Despite having different collection methodologies, these pre-training datasets are highly correlated in terms of how much information they contain related to different QA pairs.}   \label{tab:rank_correlation}
\end{table}

\subsection{Can We Scale Up Models?}\label{subsec:scaling}
Using larger models consistently produces better QA performance. However, our results suggest that one would need immensely large LMs to achieve high accuracy on long-tail questions. In Figure~\ref{fig:bloom_nq_scale} we plot a scaling trend line for rare fact learning, where we show BLOOM accuracy on rare instances from Natural Questions ($<100$ relevant docs) as a function of the log of the model size. The empirical log-linear trend---which approximates the scaling extremely well ($R^2 = 0.98$)---shows that in order to match a strong supervised baseline \cite{izacard2020distilling} or human performance, one would need a BLOOM model with over $10^{18}$ (one quintillion) parameters.\footnote{For this experiment, the supervised and human accuracies are computed over the validation set whereas the scaling trend is computed using the train and validation sets.}
We see similar trends for other models and datasets (see Figure~\ref{fig:scaling_appendix} in the Appendix).

\paragraph{Modifying the Training Objective} Another option similar to scaling up models is to directly modify the training objective to encourage memorization. One simple method to accomplish this is to increase the number of training epochs. All of the LMs that we study do limited epochs, as it is generally seen as preferable to use large enough pre-training datasets so that the LM completes one epoch of training when the compute budget is exhausted~\cite{raffel2020exploring}. However, in the context of QA, it may be preferable to increase epochs and reduce data size to ensure models memorize as much as possible. Alternatively, one could consider modifying the training loss to encourage the model to focus on salient facts~\cite{guu2020retrieval} or designing a curriculum to minimize forgetting~\cite{jagielski2022measuring}.

\begin{figure}[t]
\centering
\includegraphics[trim={0.0cm 0.0cm 0.0cm 0.0cm},clip,width=\linewidth]{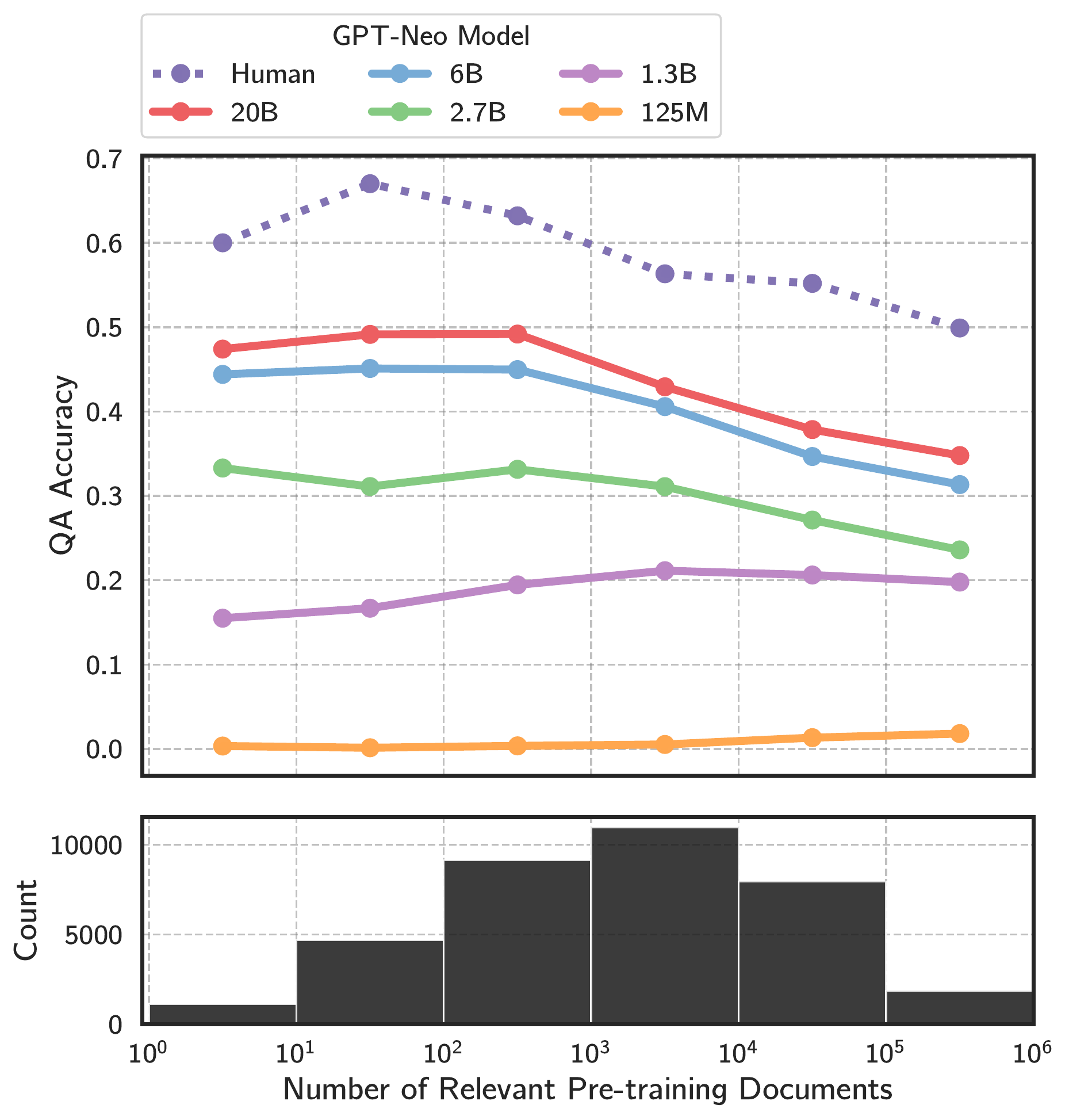}
\vspace{-0.7cm}
\caption{Models with access to the required background context do not struggle on questions with low relevant document count. Concretely, we provide questions and gold paragraphs to GPT-Neo models on Natural Questions, and their accuracy trends roughly match the trends of humans.}
\label{fig:gold_augmented_fact_count}
\end{figure}

\begin{figure*}[t]
\centering
\begin{minipage}{.48\textwidth}
    \centering
    \includegraphics[trim={0.0cm 0.0cm 0.0cm 0.0cm},clip,width=\linewidth]{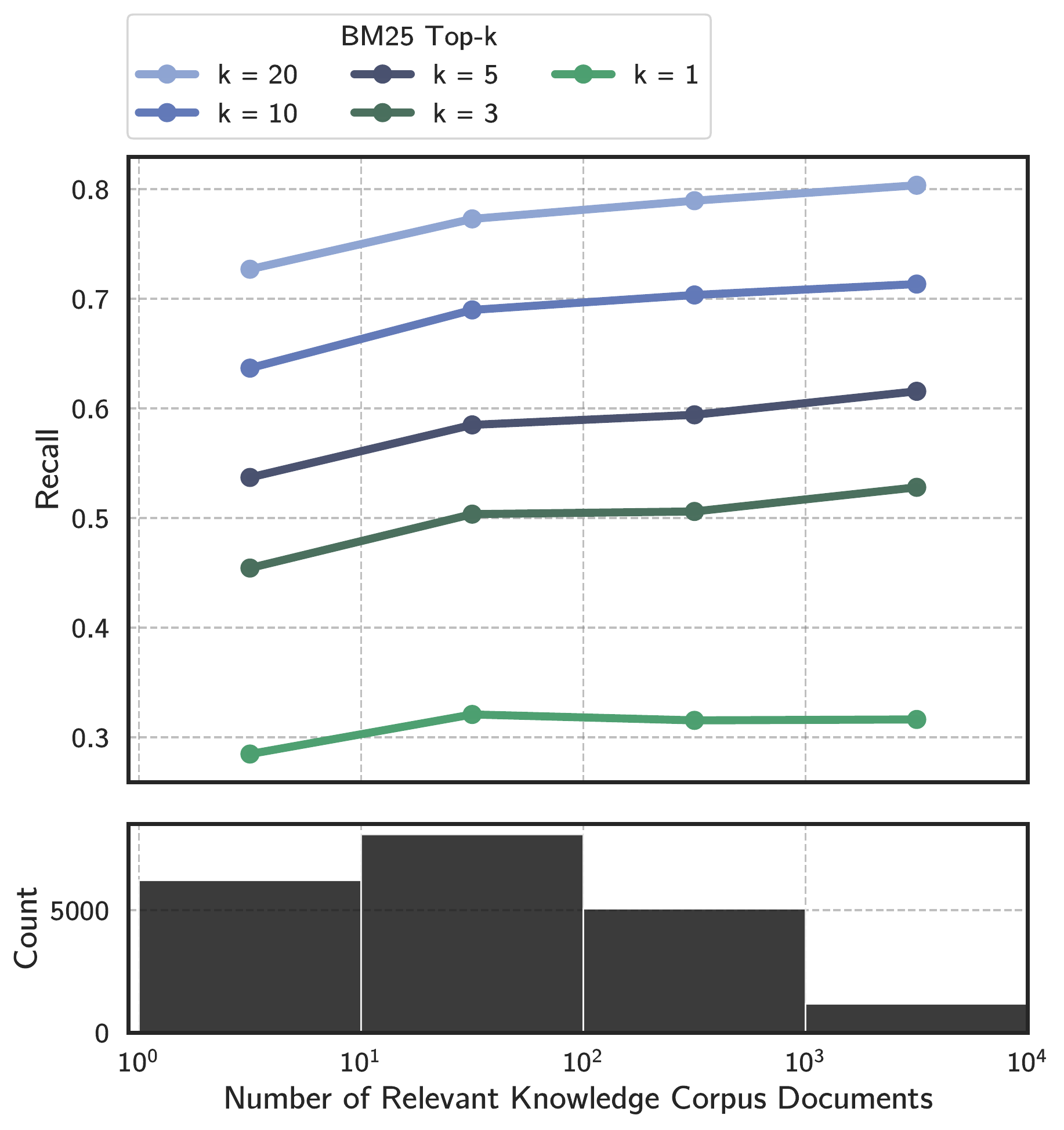}
    \caption{Retrieval systems such as BM25 have a mild dependence on document count. Above we plot the top-$k$ recall for BM25 on Natural Questions for different values of $k$.}
    \label{fig:bm25_recall}
\end{minipage}%
\hfill
\begin{minipage}{.48\textwidth}
    \centering
    \includegraphics[trim={0.0cm 0.0cm 0.0cm 0.0cm},clip,width=\linewidth]{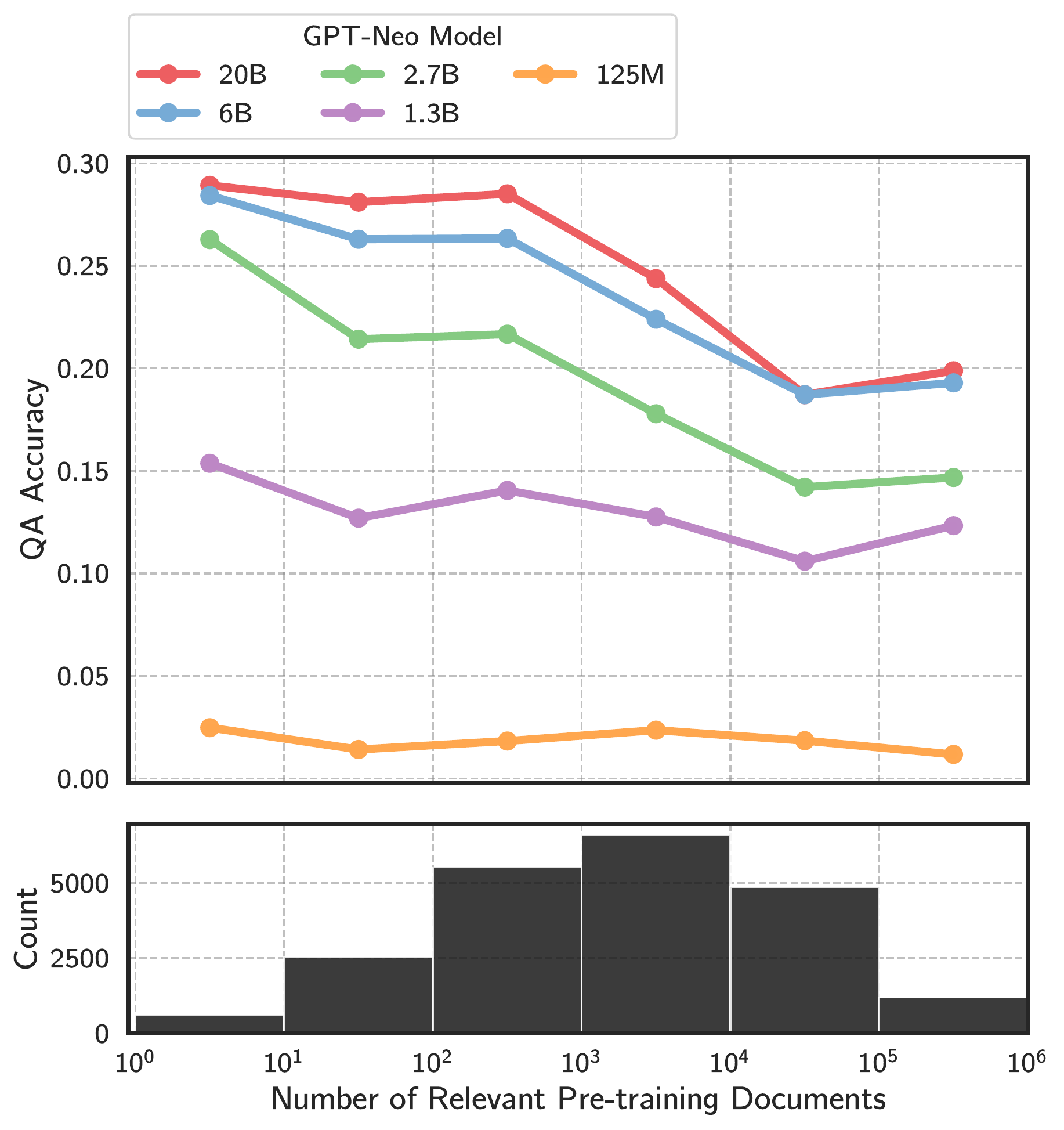}
    \caption{Retrieval-augmented LMs no longer exhibit low accuracy on rare instances. We plot GPT-Neo accuracy on Natural Questions when augmented with three paragraphs from BM25.}
    \label{fig:bm25_augmented_fact_count}
\end{minipage}
\end{figure*}

\subsection{Can We Use Retrieval Augmentation?}\label{subsec:retrieval}

Thus far, we use LMs as isolated systems that do not leverage external information. However, for knowledge-intensive tasks, a natural alternative is to make LMs \emph{retrieval-augmented}, i.e., combine them with a retrieval module that returns relevant textual contexts~\cite{lewis2020retrieval,guu2020retrieval,karpukhin2020dense}. Here, we study whether retrieval-augmented models can mitigate the dependence on the amount of relevant knowledge in the pre-training data.

\paragraph{Oracle Retrieval} We first study an oracle setting where we provide LMs with a gold paragraph from Wikipedia that supports each answer in Natural Questions~\cite{petroni2020context}. We use the 300-word segment that surrounds the ground-truth answer from the gold Wikipedia page and evaluate the 2-shot accuracy of GPT-Neo.
Figure~\ref{fig:gold_augmented_fact_count} shows that oracle retrieval-augmentation dramatically boosts accuracy over closed-book models, especially on rarer instances. Similar to \citet{liu2021challenges}, we also find that QA accuracy actually goes \emph{down} as the number of relevant documents increases---the opposite trend of closed-book LLMs. As discussed in Section~\ref{subsec:correlation}, humans exhibit the same trend, likely because rare questions are easier on average when relevant context information.

\paragraph{BM25 Retrieval} We next follow a common retrieval-augmented baseline, where we use a BM25 retriever~\cite{bm25} to select paragraphs from Wikipedia. We add the top-3 highest scoring paragraphs into the prompt for both the in-context training examples and the test question.
We verify that at least one of the retrieved paragraphs contains the answer for each in-context training example, to ensure that the LM learns to utilize on the documents.

We first evaluate the BM25 retriever's top-$k$ recall on its knowledge corpus (Wikipedia) as a function of relevant document count, and plot the results in Figure~\ref{fig:bm25_recall}. We find that BM25 attains reasonably high recall, especially for larger values of $k$. However, the BM25 retriever still shows a mild dependence on relevant document count.
We next evaluate the accuracy of BM25-augmented GPT-Neo models on Natural Questions and plot the results in Figure~\ref{fig:bm25_augmented_fact_count}.
Overall, retrieval-augmented models outperform their closed-book counterparts across all ranges of relevant document counts, and especially on rare examples.
These results suggest that retrieval augmentation provides a promising path towards improving performance on questions with few relevant documents in the pre-training dataset.
\section{Related Work}

\paragraph{Identifying The Origins of Few-shot Learning} Our work contributes to an emerging line of research that explains the success of zero- and few-shot learning in language models by tracing their behavior back to the pre-training data. For example,  \citet{razeghi2022impact} show mathematical reasoning capabilities can be correlated with training data frequency, and \citet{shin2022effect} and \citet{han2022orca} show that training corpus source can influence few-shot accuracies.

The most similar work to ours in this context is \citet{elazar2022measuring}, who use causal inference to measure the effect of pre-training data statistics on QA performance.
Their main focus is testing the extent to which LMs answer questions using heuristics based on co-occurrences between subjects, objects, and textual patterns in the pre-training data.
Our main focus is to measure the relationship between the knowledge learned by an LLM and the prevalence of that knowledge in the pre-training data.
Moreover, we also conduct re-training experiments and study how model scaling and retrieval-augmentation affect knowledge learning.

\paragraph{Memorization and Privacy} Past work studies training data memorization from the perspective of privacy, i.e., how LMs inadvertently reveal private text~\cite{carlini2019secret,carlini2021extracting,lee2021deduplicating}. These works focus on how LMs memorize and repeat \textit{verbatim} text samples, and the effect of duplicating those texts in the training set~\cite{Kandpal2022Deduplicating}. 
Doing so has various limitations, as memorization can be harmful or beneficial even in non-verbatim cases~\cite{ippolito2022preventing}.
Our work takes studies non-verbatim memorization in the context of QA---our LMs memorize facts in text form and then answers questions about those facts at test time.

\paragraph{Memorization and Fact Learning} Existing work also analyzes the relationship between the pre-training data and the factual knowledge of LLMs. 
\citet{akyurek2022tracing} look to automatically identify which documents were most influential for a language model's QA predictions. Our work instead directly identifies and estimates the number of relevant documents via entity linking large corpora. Other work notices a correspondence between model accuracy and data frequency for different knowledge-intensive tasks~\cite{petroni2019language,kassner2020pretrained,de2020autoregressive,wei2021frequency,fevry2020entities} and for domains outside of NLP~\cite{pmlr-v139-rao21a}. Our paper reports similar findings, but scales this analysis to massive LM pre-training datasets and model sizes.

In concurrent and independent work, \citet{mallen2022not} study how QA performance correlates with frequency in the pre-training data. Unlike our work, they do not use entity linking methods to count occurrences and instead use proxies such as entity popularity on Wikipedia. They also find QA accuracy is highly correlated with  pre-training data frequency and show that retrieval models can improve long-tail knowledge. Our work differs in that we conduct causal re-training experiments and find that model scaling is highly beneficial to long-tail QA performance.
\section{Conclusion and Future Work}

Large language models demonstrate impressive few-shot learning capabilities that arise from simply training on large-scale internet text. With the open-source release of LLMs---and their associated pre-training datasets---the research community can now begin to understand the origins of these capabilities. Our work is one of the first to relate an observed phenomenon in LLMs back to the pre-training data itself.
In our case, our results are negative: while LLMs achieve moderate performance on open-domain QA benchmarks, they are mainly successful on questions that probe knowledge that appears widely in their pre-training datasets. 

Our work raises numerous directions for further inquiry, namely, how to improve retention of long-tail knowledge given that simply scaling up model and dataset size will likely be insufficient. We are personally excited about improving retrieval-augmented LMs, especially with regards to their efficiency and retrieval accuracy.
Moreover, our work focuses on knowledge learning as it relates to factoid question answering, but we leave open the question as to whether similar relationships exist for other types of tasks, be it knowledge-intensive or otherwise.
Relatedly, even though our work analyzes the impact of memorization on question answering, our results may have implications for other tasks that require using (or avoiding) memorized knowledge, e.g., analyzing private text, performing commonsense reasoning, or predicting source code.
Finally, we hope that future evaluations of few-shot learning can continue to shed light into model behavior by tracing accuracy back to properties of the pre-training data. In particular, our work shows that by performing such an analysis, one can help elucidate the successes and failures of existing models, as well as help to identify possible paths forward to improve today's systems.
\section*{Acknowledgements}

We thank Sewon Min, Sameer Singh, Katherine Lee, and the members of UNC NLP for their valuable feedback. Eric Wallace is supported by the Apple Scholars in AI/ML Fellowship. This work was supported by NSF-AI Engage Institute DRL-2112635.

\bibliography{journal-abbrv,references}
\bibliographystyle{icml2023}

\clearpage
\appendix
\onecolumn


\section{Additional Results: Relevant Document Scaling}\label{appendix:fact_scaling}

\begin{figure}[!htb]
\centering
    \includegraphics[width=0.48\textwidth]{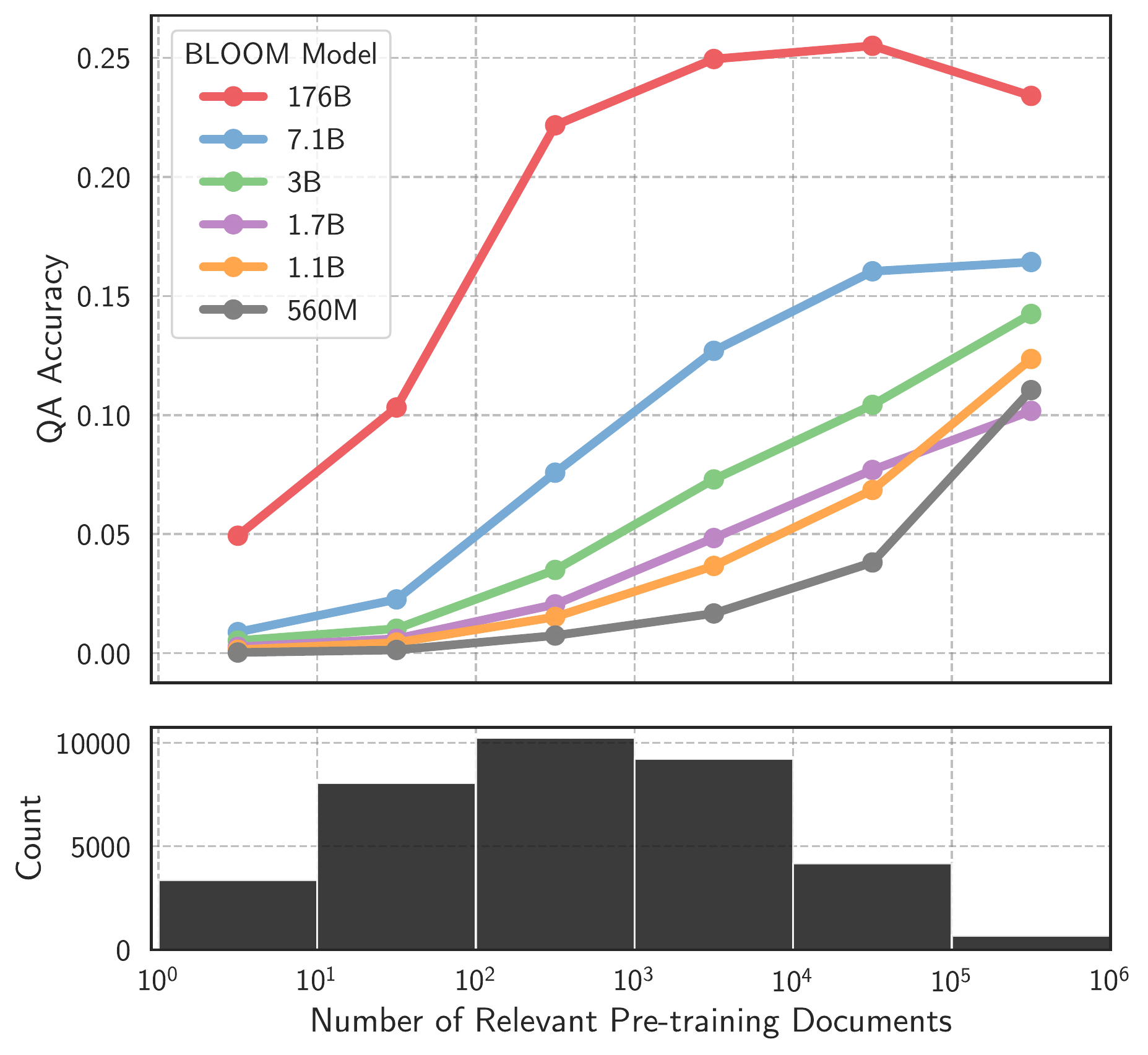}
    \label{fig:bloom_neo_nq}
\vspace{-0.3cm}
\caption{We show results for Natural Questions for BLOOM. The trends match those seen in TriviaQA, although the accuracy is lower overall for Natural Questions.}
\label{fig:nq}
\end{figure}

\begin{figure}[!htb]
\centering
\subfigure[]{
    \includegraphics[width=0.4\textwidth]{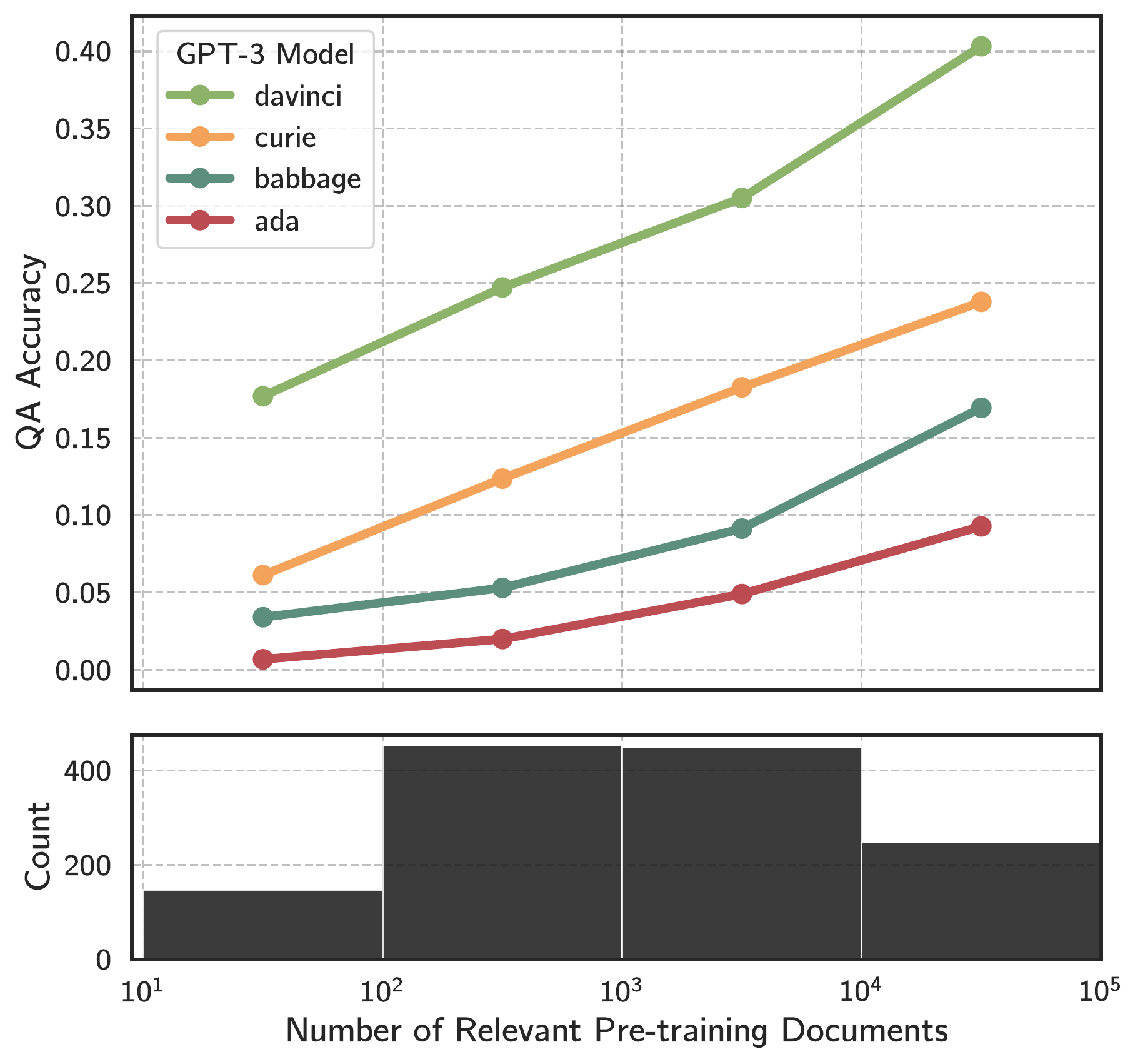}
    \label{fig:gpt3_nq}
    }
\subfigure[]{
\includegraphics[width=0.4\textwidth]{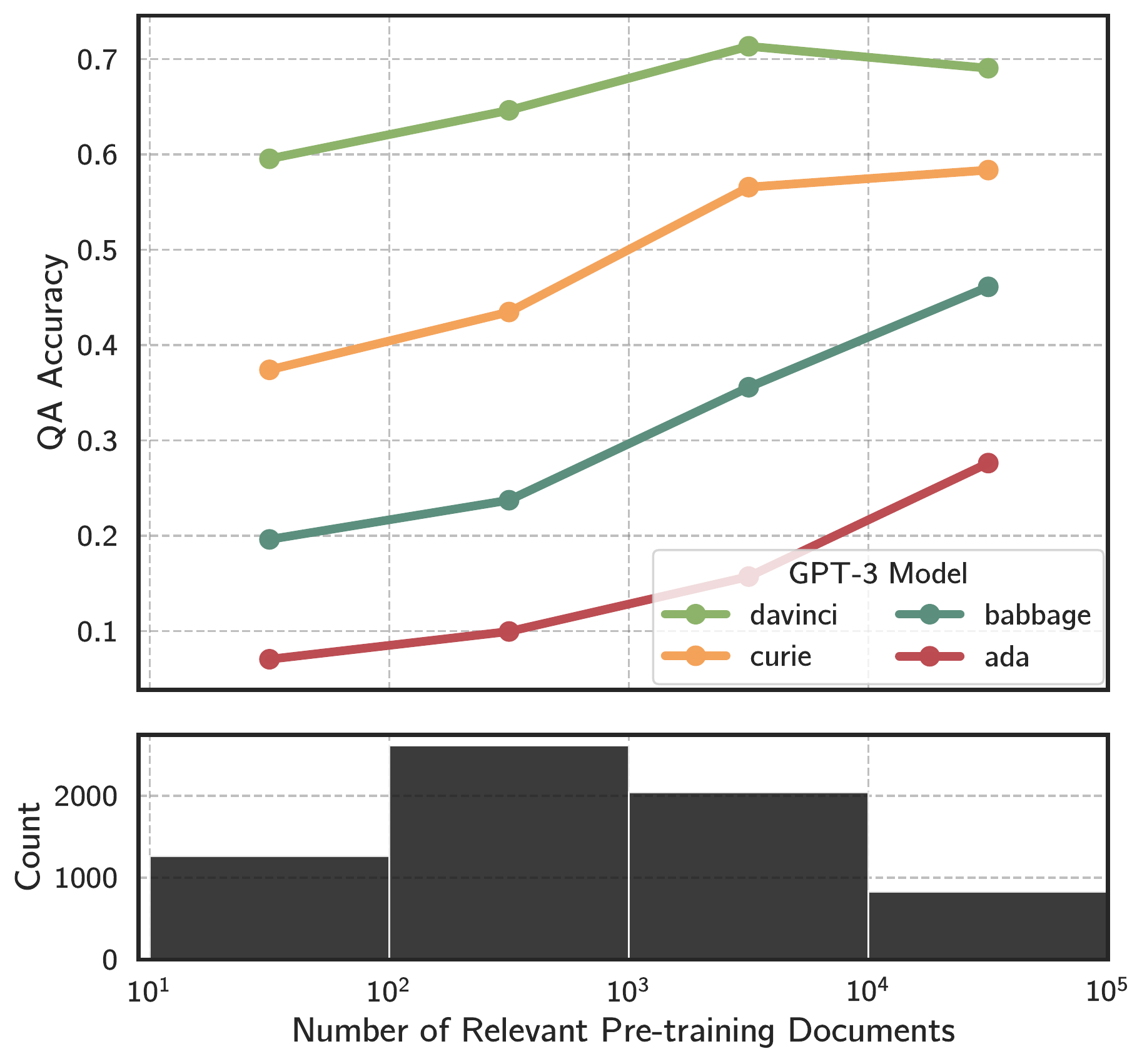}
  \label{fig:gpt3_tqa}
} 
\vspace{-0.3cm}
\caption{We present the QA results for GPT-3, with Natural Questions shown in \textbf{\subref{fig:gpt3_nq}} and TriviaQA shown in \textbf{\subref{fig:gpt3_tqa}}. The trends match those seen in BLOOM and GPT-Neo. Note that our estimates for the number of relevant pre-training documents may be inaccurate because the training data for GPT-3 is not public.}
\label{fig:gpt3}
\end{figure}

Here we show how QA performance is related to the number of relevant pre-training documents for the BLOOM on Natural Questions (Figure~\ref{fig:nq}) and the GPT-3 model family on TriviaQA and Natural Questions (Figure~\ref{fig:gpt3}). Like the results in the main text, models are significantly better at answering questions about facts that are well supported in the pre-training data and model scale improves knowledge acquisition. 

Note that our estimates for the number of relevant pre-training documents for the GPT-3 model family may be inaccurate since the training data for GPT-3 is not public. Instead, we estimate these relevant document counts using the open source dataset OpenWebText, which was collected with a similar process to the reported collection methodology for the GPT-3 pre-training dataset. 


\section{Additional Results: Model Scaling}\label{appendix:model_scaling}
\begin{figure*}[!htb]
\centering
\subfigure[]{
    \includegraphics[width=0.31\textwidth]{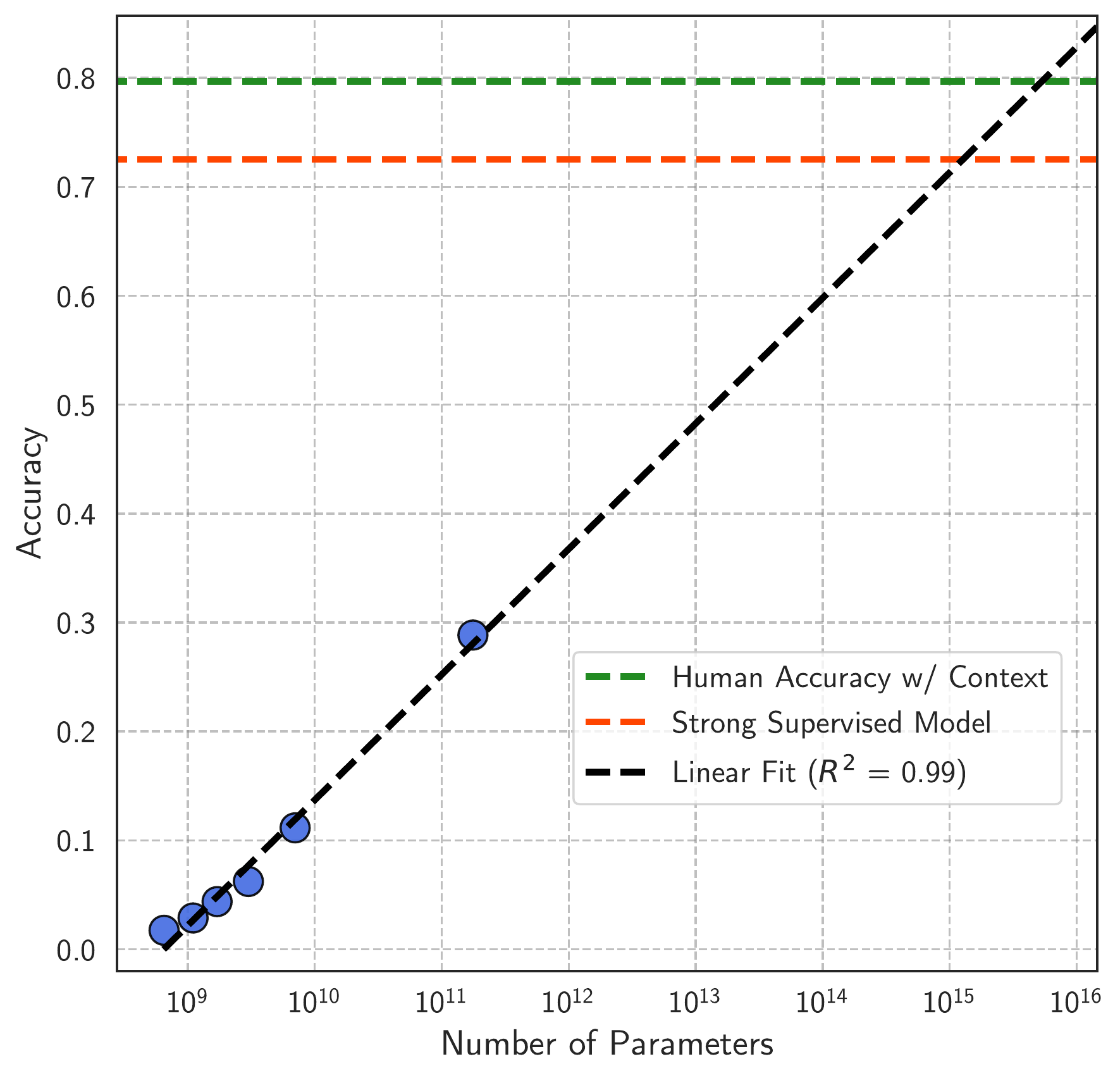}
    \label{fig:bloom_tqa_scale}
    }
\subfigure[]{
\includegraphics[width=0.31\textwidth]{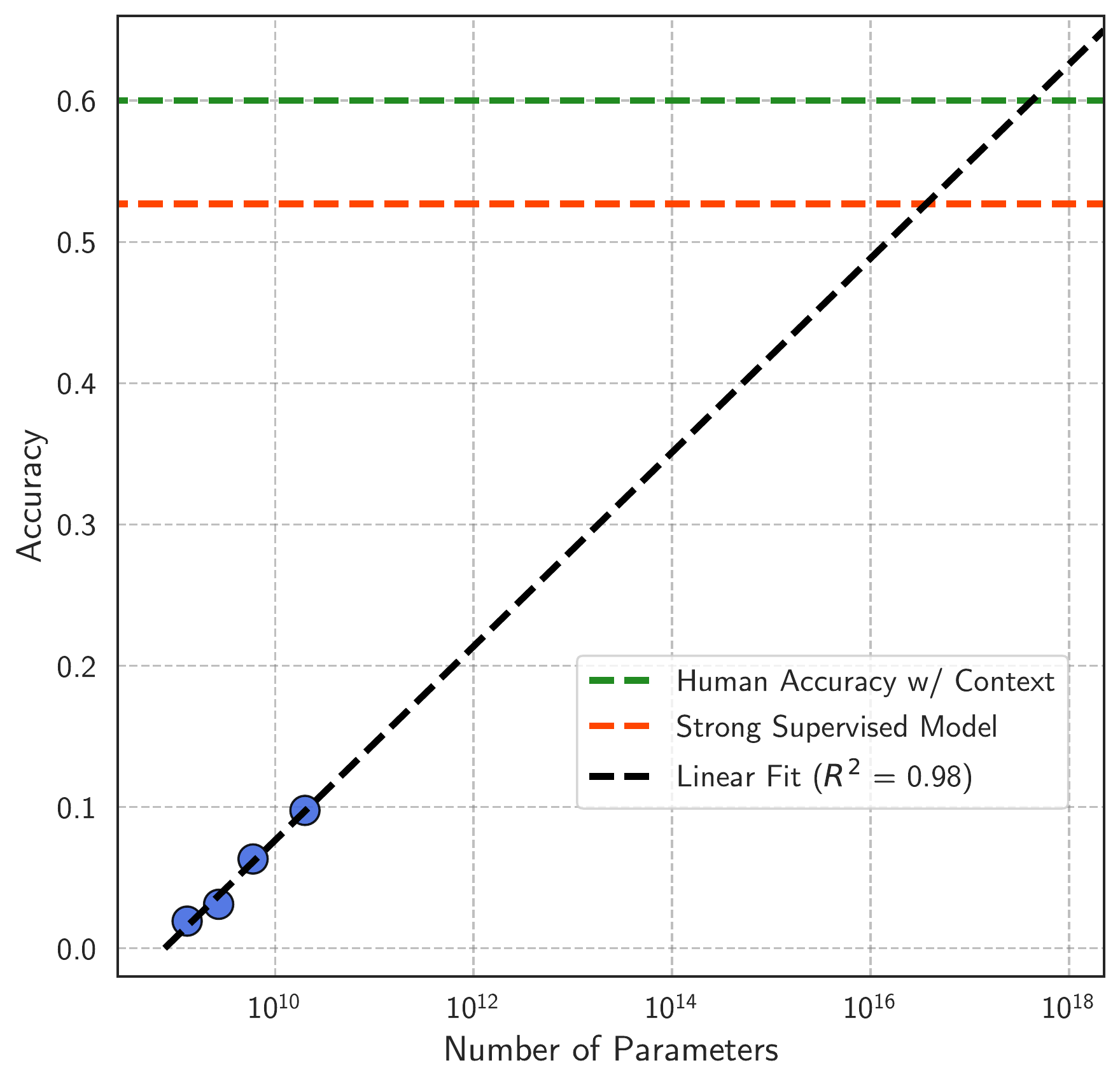}
  \label{fig:gpt_neo_nq_scale}
}
\subfigure[]{
\includegraphics[width=0.31\textwidth]{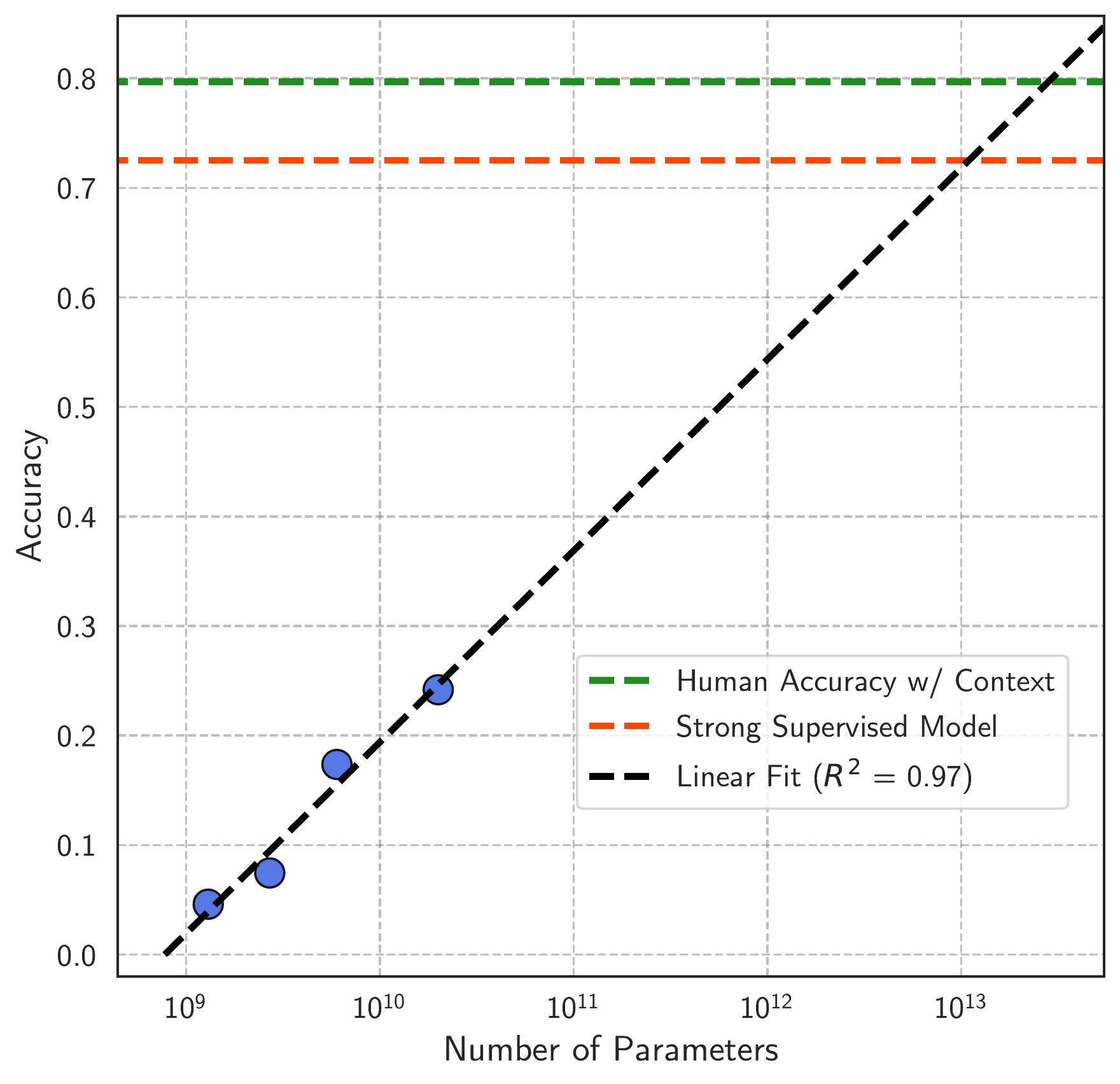}
  \label{fig:gpt_neo_tqa_scale}
}
\vspace{-0.3cm}
\caption{We present additional scaling laws for BLOOM on TriviaQA \textbf{\subref{fig:bloom_tqa_scale}}, and GPT-Neo on Natural Questions \textbf{\subref{fig:gpt_neo_nq_scale}} and TriviaQA \textbf{\subref{fig:gpt_neo_tqa_scale}}. All the trends are similar---we will need to scale up models dramatically to reach high QA accuracy---but the exact degree to how much we would need to scale models changes across the different settings.}
\label{fig:scaling_appendix}
\end{figure*}

In this section we show additional results for how long-tail QA accuracy scales with model size for the BLOOM model family on TriviaQA and the GPT-Neo model family on Natural Questions and TriviaQA (Figure~\ref{fig:scaling_appendix}). The log-linear trend matches the results shown in the main text.

\section{Relevant Document Counting Heuristics}\label{appendix:counting}
\begin{figure*}[!htb]
\centering
\subfigure[]{
    \includegraphics[width=0.4\textwidth]{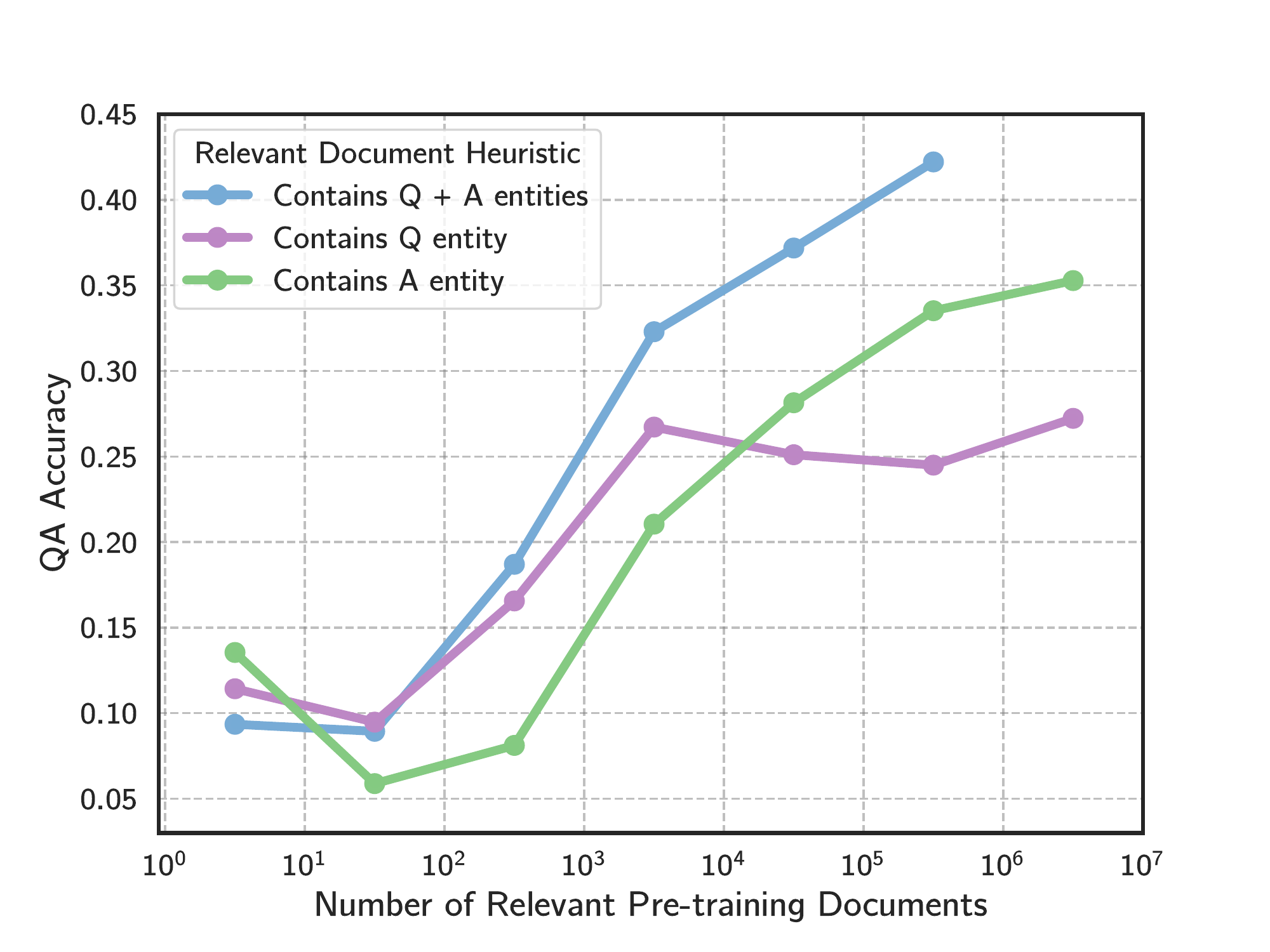}
    \label{fig:heuristics}
    }
\subfigure[]{
\includegraphics[width=0.4\textwidth]{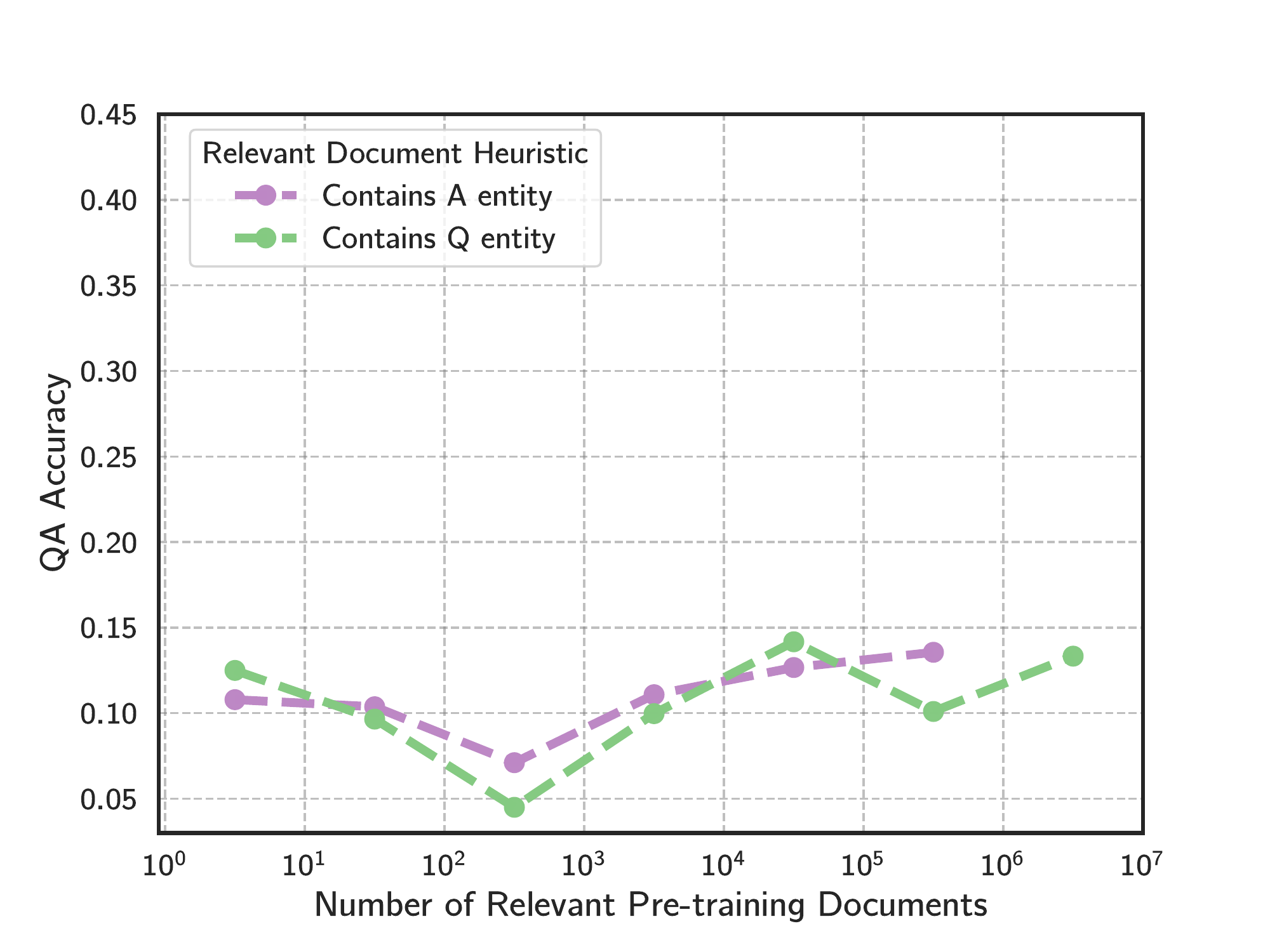}
  \label{fig:co_occurrence_control}
} 
\vspace{-0.3cm}
\caption{In \textbf{\subref{fig:heuristics}}, we plot the relationship between model accuracy and the count of the question entity alone, as well as the answer entity alone. QA accuracy increases as both of these counts increase. In \textbf{\subref{fig:co_occurrence_control}}, we consider only f QA pairs with few question and answer entity co-occurrences ($<5$ documents). For this subpopulation of QA pairs, neither of the baseline heuristics are correlated with QA accuracy.}
\label{fig:question_answer_only}
\end{figure*}

In this section, we analyze the difference between our relevant document heuristic, which counts documents where the salient question and answer entity co-occur, compared to two simple baselines: counting documents containing the question entity and documents containing the answer entity. In Figure~\ref{fig:heuristics} we show that all three document counting heuristics are correlated with QA accuracy. However, as seen in Figure~\ref{fig:co_occurrence_control} the correlation of the two baseline counting methods with QA accuracy disappears when only considering QA examples where the question and answer entity co-occur few ($<5$) times in the pre-training data. Thus, these baseline counting heuristics appear correlated with QA accuracy simply because they are simply correlated with question and answer entity co-occurrence (i.e., common entities tend to co-occur with other entities more frequently) rather than causally related to QA performance.

\end{document}